\documentclass{article}

\usepackage{arxiv}

\usepackage[utf8]{inputenc} 
\usepackage[T1]{fontenc}    
\usepackage{hyperref}       
\usepackage{url}            
\usepackage{booktabs}       
\usepackage{amsfonts}       
\usepackage{nicefrac}       
\usepackage{microtype}      
\usepackage{lipsum}
\usepackage{graphicx}
\graphicspath{ {./images/} }
\usepackage{setspace}
\usepackage{natbib}
\newcommand{\citeg}[1]{\citep[e.g.,][]{#1}}
\usepackage{doi}

\title{A perishable ability? The future of writing in the face of generative artificial intelligence}

\author{
 Evandro L. T. P. Cunha \\
  Universidade Federal de Minas Gerais (UFMG)\\
  Belo Horizonte, MG, Brazil \\
  \texttt{cunhae@ufmg.br} \\
}

\begin{document}
\maketitle
\begin{abstract}
The 2020s have been witnessing a very significant advance in the development of generative artificial intelligence tools, including text generation systems based on large language models. These tools have been increasingly used to generate texts in the most diverse domains – from technical texts to literary texts –, which might eventually lead to a lower volume of written text production by humans. This article discusses the possibility of a future in which human beings will have lost or significantly decreased their ability to write due to the outsourcing of this activity to machines. This possibility parallels the loss of the ability to write in other moments of human history, such as during the so-called Greek Dark Ages (approx. 1200 BCE - 800 BCE).
\end{abstract}


\section{Introduction}
Writing is arguably the most powerful and transformative technology that humans have ever developed. \citet{blatz1944education} considers that the invention of this tool made it possible to permanently record, for succeeding generations, the accumulation of human experience. According to \citet{dellongo2014writing}, the development of writing has enabled human language to cross the boundaries of time, space, and modality thanks to its capability to store and retrieve information. In fact, writing is sometimes regarded with such amazement that many societies suppose that it came from heaven \citep{schmandt1986origins} – which shows not only its power, but also the lack of understanding of its progressive development as a technological tool. From an economic point of view, the adoption of written records was essential to the emergence of cities and states, and facilitated trade and bureaucratic administration, in addition to having stimulated the advancement of philosophy and science \citep{hanson2004superorganic}. It is not by chance, therefore, that, historiographically, the invention of writing is what marks the passage from what we call \textit{prehistory} to what we consider \textit{history} \citep{goody1963literacy}.\footnote{Even though this division is questionable, in particular due to the existence of contemporary societies that have not adopted writing. For a discussion on this matter from an archaeological perspective, see \citet{fredericksen2000history}.}

In the first decades of the 21st century, the relationship between humans and writing has been changing dramatically. Email, instant messaging and social media, among other text-based computer-mediated communication tools, have provided new forms of textual interaction and contact with written texts, in addition to having promoted the appearance of new text genres \citep{giltrow2009genres}. In many languages, the use of these communicative tools has fueled the rise of orthographically unconventional forms, that perform as shortcuts to reduce writing time and to achieve the brevity required by character limits, as well as being innovations that mark the language use by young people \citep{verheijen2013messaging}. As at various times in history, changes in people's linguistic and orthographic behavior generate criticism and feed the idea that "the language is dying".\footnote{For an example from the Late Antiquity period, see the Appendix Probi, that lists Latin forms considered wrong, thus to be avoided, and their corrections \citep{quirk2017hypercorrection}.} From a strictly scientific point of view, however, this idea makes little sense: language change is a natural process that happens at every stage of every living language, and even though written language is more conservative than spoken language, orthographic changes take place whenever there is writing.

More recently, technological advances in the field of artificial intelligence have created tools capable of paraphrasing, summarizing and even generating texts from scratch. These natural language generation tools have become more and more popular in the early 2020s, when general-domain chatbots based on large language models were launched, with great media interest. The highly satisfactory performance of these text generators for many day-to-day tasks has led people to use them for producing the most diverse types of text, from academic recommendation letters \citep{kaplan2024bias} to poetry translations \citep{gao2024poetrytranslation}.

The medium and long-term effects of the heavy use of tools based on generative artificial intelligence\footnote{Generative artificial intelligence "is a form of AI that can autonomously generate new content, such as text, images, audio, and video" \citep[p. 208]{lv2023generative}.} are still not very clear. In this article, it is argued that, although it may seem exaggerated and even absurd to some, it should not be excessive to consider that the human capacity to produce written texts may decrease considerably or even disappear in the foreseeable future due to the increasingly accelerated development and use of artificial intelligence tools for text generation. With the possibility of an ever-decreasing need to produce their own written texts, humans could then gradually abandon the practice of writing, thus converting it into an activity to be performed exclusively or almost exclusively by machines – similarly to other activities that have already gone through or are going through this process.

\section{Writing as a nonperennial technology: the case of Linear B}
It is not news that, over the course of human history, technologies come and go. Different models of adoption, continued usage and abandonment of technologies have been proposed, and studies focused on the acceptance and renunciation of technologies in fields such as education \citeg{granic2019tam}, healthcare \citeg{nadal2020mhealth} and communication \citeg{birnholtz2010adopt} were carried out.

Nonetheless, some technologies have been around for so long that they feel eternal. This is probably the case with writing, a technology that was first attested in Southern Mesopotamia around 3500 BCE\footnote{Yet it is now widely accepted that writing was developed independently also in East Asia and Mesoamerica \citep{fischer2003writing}.}, based on previous forms of representing language and as a means of record-keeping through signs and pictograms marked on rocks and clay tokens. Over the centuries, writing established itself as a powerful technology and spread across the globe, evolving into a robust tool for communication, literature, law and philosophy. It is indisputable that its dissemination facilitated the exchange of ideas, the preservation of knowledge and, as a consequence, the formation of complex societies.

Despite all the power contained in writing, however, this technology cannot be considered free of the risk of being cast aside by a civilization. The abdication of writing systems throughout human history is, of course, not as common as their adoption, but there might have been a few instances.\footnote{Two possible examples of this are the Rongorongo script of Easter Island \citep{fischer1997rongorongo} and the Indus Valley script \citep{fairservis1983indus}, which remain undeciphered and might have been abandoned due to societal changes, such as environmental or cultural shifts.} The most significant example of this phenomenon is probably the abandonment of Linear B, in the beginning of the so-called Greek Dark Ages.

Linear B is a writing system developed during the Late Bronze Age in order to write the Mycenaean Greek language. It was in use between approximately 1500 BCE and 1200 BCE, and derives directly from Linear A – the writing system used to write in the Minoan language. Linear B script was deciphered in the early 1950s and its texts, present in more than five thousand surviving clay tablets, give us a broad overview of the primary economic priorities within the Mycenaean palatial territories \citep{palaima2010linearb}. Certainly, during the Late Bronze Age the activity of writing was not as widespread as it is in most modern societies. This can be observed by the very nature of the Linear B texts that have reached us, which deal especially with administrative, economic and military topics. In spite of being considered by some as a complex writing system, \citet[p. 364]{palaima2010linearb} adds that several hundred tablet writers "had no difficulty using the script effectively and routinely on a daily basis".

Around the late 12th century BCE, the Mycenaean civilization went into decline, for reasons that are still unclear.\footnote{But possibly triggered by a combination of factors such as social unrest, invasion by foreign powers, economic decline and environmental changes. For an in-depth and intricate discussion on the Late Bronze Age collapse, see, for example, \citet{drews1993bronzeage} and \citet{millek2023destruction}.} According to \citet{middleton2019collapse}, the collapse of this civilization can be identified from five main features of the archaeological record: (a) destructions at most major palace sites; (b) cessation of major building projects; (c) loss of most fine arts and crafts; (d) cessation of Linear B use; and (e) site abandonments. The 'cessation of Linear B use' feature is particularly relevant for us. \citet{middleton2019collapse} explains that:

\begin{quote}
Records had been kept in Linear B in some palaces since LM IIIA1 (Knossos) (early 14th century BCE); they were a long-lived feature of the palace operations and indicate a tradition of (probably very limited) literacy. They represent the reach of the palace and its personnel into the surrounding territory and its duties with regard to (state?) religious occasions. This tradition, going back around two centuries, ended in the collapse. This may indicate the fracturing of relationships between palaces and their (former) territories, or locations where their influence had been felt. The abandonment of writing must be considered deliberate as there would have been some memory of it through c. 1200 BCE. Syllabic writing, influenced by Linear B, continued on Cyprus through the LBA/EIA transition and was used to write in Greek, but in the Aegean there was no more writing until the much later adoption of the alphabet. \citep{middleton2019collapse}
\end{quote}

Linear B was primarily used for administrative and economic purposes in the Mycenaean palace centers. Hence, with the collapse of these centers and the societal upheaval that followed, the use of the script declined and it eventually fell out of use, since there was no longer a need or system in place to maintain it. \citet{middleton2019collapse} even concludes, due to the abrupt disappearance of archaeological records, that the abandonment of writing in this context was deliberate. Also, it appears plausible to raise the hypothesis that, during the challenging period that ended up being known as the Greek Dark Ages, writing may have been deemed superfluous and dispensable for those circumstances, making it a useless technology in the eyes of that society.

\section{The possibility of erosion of human writing abilities}
It is perhaps unlikely that modern societies will face the same challenges that Mycenaeans went through in the Late Bronze Age. Despite this, there are still several scenarios for the possible downfall of humanity, including the threat of near-extinction due to nuclear war \citep{ehrlich1983nuclearwar}, pandemics \citep{peters2022pandemic} or global warming \citep{carpenter2009extinction}. Also, some scholars have raised the possibility of the emergence of major crises due to the development of artificial general intelligence devices: there are concerns, for example, that these systems could surpass human intelligence and become uncontrollable or act against human interests \citep{growiec2024existentialrisk}. These situations could eventually lead to a scenario in which writing becomes superfluous in the face of other difficulties to be experienced by the remaining human beings.

Even if none of these catastrophic scenarios materialize and humanity flourishes more and more, a risk to the maintenance of the human ability to write is the expansion of the use of tools that make this knowledge irrelevant. The reduction in the need to use the competence of writing could lead to a situation in which knowing it is unnecessary for a large part of the population that currently uses it. The act of writing texts could well be completely replaced by at least one of two other present technologies, or by a combination of both: speech transcription and the automatic production of texts using tools based on generative artificial intelligence. In this sense, the erosion of human writing abilities due to advanced text generators might be similar, for example, to the decline in human ability to drive cars due to to the advent of autonomous vehicles, as predicted by \citet{havlickova2020driving}: as autonomous vehicles become the norm for daily transportation, there will be less incentive for individuals to acquire or maintain proficiency in driving – and as generations grow up in a world dominated by self-driving cars, the knowledge and muscle memory associated with operating vehicles manually will probably fade into obscurity.

This scenario does not necessarily imply that human will lose the ability to read, since it is possible for someone to have passive literacy, which refers to the ability to read and understand written text but not necessarily to actively engage in writing or producing written content \citep{cressy1977illiteracy}. Individuals with passive literacy skills can comprehend written materials, but may not frequently engage in writing tasks themselves. This ability can still be valuable for consuming information, learning and communication, even if the individual does not actively participate in writing activities.

It also seems reasonable to think of a scenario in which the ability to write is maintained only by professional writers, especially those associated with creative writing. This scenario is in line with early uses of writing, when it was a specialized skill reserved for a small, educated elite class \citep{blair2021scribes}. Still, this is probably not the case for content creators, technical writers and journalists, who are already being replaced by artificial intelligent text generators \citep{miroshnichenko2018ai}.

\section{ChatGPT’s opinion on this matter}
As of 2024, ChatGPT\footnote{\url{https://chat.openai.com/}}, developed by OpenAI, is probably the most well-known generative artificial intelligence tool, acting as a chatbot and virtual assistant that interacts with users following their questions and commands – known as \textit{prompts}. The usage of ChatGPT in different scenarios has been the subject of a myriad of research projects, interested in aspects such as its mathematical capabilities \citep{frieder2023mathchatgpt}, its effectiveness to provide feedback to students \citep{wang2024argumentation} and the possibilities of its use for language teaching and learning \citep{kohnke2023chatgpt}, among several others.

When asked about the possibilities of the risk of erosion of human writing skills due to generative artificial intelligence, ChatGPT (GPT-3.5-based version) replied as follows\footnote{All the interactions presented in this article were held on 4 September 2023.}:

\begin{quote}
\textbf{Prompt:} Hi! Do you think that, in the future, humans will lose their writing abilities due to generative artificial intelligence?

\textbf{ChatGPT’s answer:} It's unlikely that humans will completely lose their writing abilities due to generative artificial intelligence (AI) in the foreseeable future. While AI, like the one you are currently interacting with, has made significant advancements in generating text and can assist with writing tasks, there are several reasons why humans will continue to value and use their writing skills:

1. Creativity and originality: AI can generate text based on patterns and data it has learned, but it lacks true creativity and the ability to come up with entirely original ideas and perspectives. Human writers can bring unique creativity, emotions, and personal experiences to their writing that AI cannot replicate.

2. Critical thinking and analysis: Writing often involves the expression of complex thoughts, arguments, and analysis of various topics. Humans possess critical thinking skills that allow them to evaluate information, synthesize ideas, and present them coherently, which is challenging for AI to achieve comprehensively.

3. Emotional connection: Human writing can evoke emotions, empathy, and relatability. Personal stories, memoirs, and other forms of human expression can establish deep emotional connections with readers, which is a unique aspect of human writing.

4. Context and nuance: Understanding context, cultural nuances, and social dynamics is crucial in effective communication through writing. AI may struggle with these subtleties, leading to misinterpretations or insensitive responses in some situations.

5. Varied writing styles and purposes: Writing encompasses a wide range of styles and purposes, from poetry and literature to scientific research and legal documents. While AI can assist with certain tasks, humans will continue to specialize in various writing niches and adapt to evolving communication needs.

6. Learning and education: Writing is a fundamental skill taught in educational systems worldwide. Humans will still need to learn how to write effectively to communicate and succeed in various aspects of life, including academia, professional careers, and personal communication.

While AI can be a valuable tool to enhance and streamline the writing process, it is more likely to complement human writing abilities rather than replace them entirely. As technology evolves, it's essential for individuals to continue developing their writing skills alongside their understanding of AI's capabilities to harness the full potential of both human and artificial intelligence in various contexts.
\end{quote}

I would not expect another answer. Anecdotally\footnote{Up to the time of writing this article, I have not been able to find any work that has analyzed this specific phenomenon.}, I have observed that ChatGPT responses are systematically biased towards the view that artificial intelligence systems are relatively safe and potentially beneficial to humanity, in particular when used in conjunction with human intelligence. However, it is interesting to note that, right at the beginning of its response, the tool stated that "it's unlikely that humans will \textit{completely} lose their writing abilities" – but what about \textit{partially} lose them?

In any case, I decided to explore some of the arguments given by ChatGPT as "reasons why humans will continue to value and use their writing skills". Of course, generative artificial intelligence systems create text based on patterns in the training data without truly understanding its meaning – which is why they have been nicknamed "stochastic parrots" by \citet{bender2021dangers}. This also means that they reinforce and amplify biases and harms, especially towards marginalized populations. But it is undeniable that texts theoretically loaded with feelings can be satisfactorily generated by these tools, as we can see below:

\begin{quote}
\textbf{Prompt:} Could you write a two paragraphs love letter to Lorenza?

\textbf{ChatGPT’s answer:} My dearest Lorenza,

Every moment with you feels like a cherished dream, and every glance into your eyes reveals a world of beauty and wonder that I am endlessly grateful to explore. Your smile is my sunshine, your laughter my favorite melody, and your presence, my sanctuary. With every beat of my heart, I am reminded of the incredible fortune that brought you into my life.

In your embrace, I have found a home, and in your love, I have discovered a profound sense of belonging. You are the missing piece that has made my life complete, and I vow to cherish and adore you for all eternity. With you, Lorenza, I have found not only love but also the truest and most profound sense of happiness. You are my heart's desire, my soul's greatest joy, and I will love you today, tomorrow, and for all the days that follow.

Forever yours, [Your Name]
\end{quote}

It goes without saying that ChatGPT does not really feel anything for Lorenza. But, in practice, it does not really matter: someone interested in sending a love letter can simply ask ChatGPT to write it, and the loved one is unlikely to figure out the trick by themself. So, after the previous response, I decided to ask the tool another question:

\begin{quote}
\textbf{Prompt:} You just wrote a love letter that reads like human-written text. How can you say that your text lacks "creativity and originality" and "emotional connection", like you said before?

\textbf{ChatGPT’s answer:} I appreciate your kind words about the love letter I wrote. My previous response aimed to provide a balanced perspective on the capabilities of AI-generated text in comparison to human writing. While AI can generate text that appears creative and emotionally engaging, it's important to note that these qualities often rely on patterns and data learned from existing human-written content.

In the case of the love letter, I generated the text by drawing upon a vast dataset of human-written letters and expressions of love. The AI can mimic the style and emotional tone of such content, but it doesn't possess true emotions or personal experiences. Therefore, while it can create text that may seem emotionally connected and creative, it does so by synthesizing patterns and information from existing examples rather than generating entirely original thoughts, feelings, or experiences.

In many contexts, AI-generated text can be highly useful and valuable, but it's essential to recognize that it lacks the inherent creativity, emotions, and personal touch that are unique to human writers. AI's abilities are limited to the data it has been trained on and the patterns it has learned from that data, and it doesn't possess consciousness or subjective experiences. Human writers bring their own personal perspectives, emotions, and creativity to their work, which is something AI cannot replicate.
\end{quote}

Again, the argument goes that ChatGPT’s output is just an amalgamation of previously written texts (which is true) and that real feelings and emotions are not present (which is also very true). The answer, however, fails to recognize that the \textit{communicative intention} – i.e. "the actual message a speaker hopes to express regardless of the literal semantics of the sentences he or she employs" \citep[p. 447-448]{cheang2006humour} – is satisfied, which may be enough for a human to decide to outsource the task of writing a love letter to ChatGPT.

To conclude our conversation, I explicitly asked ChatGPT to provide arguments that could strengthen my point of view:

\begin{quote}
\textbf{Prompt:} Could you please write two paragraphs arguing that human might lose their writing skills due to generative artificial intelligence?

\textbf{ChatGPT’s answer:} While generative artificial intelligence has undoubtedly brought about significant advancements in writing assistance and automation, there is a legitimate concern that overreliance on AI tools could lead to a deterioration of human writing skills. As AI becomes more capable of generating coherent and contextually relevant text, there is a risk that individuals may opt for convenience over the cultivation of their writing abilities.

One key argument is that the convenience of AI-generated content might discourage individuals from investing time and effort in honing their writing skills. As AI tools become more accessible and user-friendly, people may opt for quick AI-generated content for various purposes, such as marketing copy, academic essays, or professional communication. Over time, this reliance on AI could lead to a decline in the practice of composing original, well-crafted pieces of writing. Additionally, as AI can provide immediate feedback and corrections, individuals might become less inclined to learn and internalize the rules of grammar and style, diminishing their ability to write independently. Thus, the ease of AI-generated content could contribute to a potential erosion of human writing skills. 
\end{quote}

I could not agree more.

\section{Conclusion}
Writing is one of the most unique human abilities and has been around for over 5,000 years. Its importance in our history is undeniable, which is perhaps why we might think of it as a skill that human beings will never abandon. In this article, though, it is argued that the human ability to operate the technology of writing is not necessarily eternal: like other technologies, it can be left out if it is no longer useful for a community. This seems to be what happened, for example, when it was abandoned, maybe deliberately \citep{middleton2019collapse}, in Greece at the end of the Late Bronze Age, around 1200 BCE, reappearing in this region only centuries later.

Although it may sound like an exaggeration to some, it seems possible to expect that the human ability to write will lose much of its importance in a foreseeable future, due to the exponential growth in the use of text generation systems based on artificial intelligence. At the current moment of technological development, the limitations of these systems are many and well documented – they include reasoning and logic failures, factual errors, bias and discrimination, among others \citep{borji2023chatgptfailures}. Specifically in the domain of academic writing, important flaws are constantly found in tests with ChatGPT and similar tools, such as the creation of fake articles when performing literature reviews \citep{haman2023litreview,kim2023editing}. However, it is not hard to conceive that, as investment in these tools continues to grow, their performance will become increasingly better and the text generated by them will be suitable for more and more situations, making them substitutes for human beings in the task of writing in several domains.

This article does not get into whether computers can think: that's not the question. The issue is that, with the amount of text already produced by humanity and the computational power to be available in the next decades, the task of generating virtually any written text tends to be outsourced to machines. This may or may not be seen as a problem – even though the benefits of writing for learning, reasoning and structuring thoughts, performing tasks and for lifelong learning are quite well proven \citep{dellongo2014writing}.

The risk of losing the ability to write, however, is small when compared to other risks linked to the fast advancement of artificial intelligence systems that many scholars have been predicting, particularly human extinction due to misaligned transformative artificial intelligence \citep{growiec2024existentialrisk}. To mitigate these possible dangers, it would be important for us to think whether we really want (and need) to run these risks. A few research areas and activities are restricted or subject to ethical, legal or regulatory constraints, such as biological/nuclear weapons research and human cloning, and the same is increasingly being applied to research on artificial general intelligence. The 2020s are a turning point in human history and the decisions that the current generation is taking will set the direction of the planet in the coming decades.

\section*{Acknowledgments}
This work was supported by Instituto Nacional de Ciência e Tecnologia em Inteligência Artificial Responsável para Linguística Computacional, Tratamento e Disseminação de Informação (INCT-TILD-IAR) (grant \# 408490/2024-1).

The fellows and mentors of the 4th Intercontinental Academia on \textit{Intelligence and AI} contributed a wide range of perspectives that enriched my reflections on this topic. I am especially thankful to Jakub Growiec for sharing his ideas on the existential risks associated with transformative AI. Any errors or omissions remain solely my responsibility.

\bibliographystyle{apalike}  
\bibliography{references}  

\end{document}